\documentclass{article}
\usepackage{ijcai26}

\usepackage{times}
\usepackage{soul}
\usepackage{url}
\usepackage[hidelinks]{hyperref}
\usepackage[utf8]{inputenc}
\usepackage[small]{caption}
\usepackage{graphicx}
\usepackage{amsmath}
\usepackage{amsthm}
\usepackage{booktabs}
\usepackage{algorithm}
\usepackage{algorithmic}
\usepackage[switch]{lineno}

\usepackage{xcolor}
\usepackage{booktabs}
\usepackage{multirow}
\usepackage{amsmath}
\usepackage{amssymb}
\usepackage{subcaption}
\usepackage{xspace}
\usepackage[table]{xcolor}

\title{EyeCue: Driver Cognitive Distraction Detection via Gaze-Empowered \\ Egocentric Video Understanding}

\author{
Lang Zhang$^1$
\and
JinYi Yoon$^{1,2}$
\and
Matthew Corbett$^3$
\and
Abhijit Sarkar$^1$
\And
Bo Ji$^1$\\
\affiliations
$^1$Virginia Tech~
$^2$Inha University~
$^3$Army Cyber Institute at West Point
\emails
{\fontsize{11pt}{11pt}\selectfont
langzhang@vt.edu,
jinyiyoon@inha.ac.kr,
matthew.corbett@westpoint.edu,
asarkar@vtti.vt.edu,
boji@vt.edu
}}

\begin{document}

\maketitle

\newcommand{\jiny}[1]{\textcolor{orange}{[Jiny: #1]}}
\newcommand{\matt}[1]{\textcolor{teal}{[matt: #1]}}
\newcommand{\lang}[1]{\textcolor{blue}{[Lang: #1]}}
\newcommand{\ky}[1]{\textcolor{cyan}{[Keyuan: #1]}}
\newcommand{\bo}[1]{\textcolor{red}{[Bo: #1]}}

\newcommand{\la}[1]{\textcolor{blue}{#1}}

\newcommand{\name}{\text{EyeCue}\xspace}

\begin{abstract}

Driver cognitive distraction is a major cause of road collisions and remains difficult to detect. Unlike manual or visual distraction, cognitive distraction is diverted by thoughts unrelated to driving, even when the driver appears visually attentive and exhibits no explicit physical movements. In this work, we propose \name, a gaze-empowered egocentric video understanding framework, to detect driver cognitive distraction. A key insight is that cognitive distraction manifests in the interaction between eye gaze and visual context. To capture this interaction, \name integrates eye gaze with egocentric video to enable context-aware modeling of the driver's attention over time. Furthermore, to tackle the limited scale and diversity of existing datasets, we introduce CogDrive, a comprehensive multi-scenario dataset that augments four existing driving datasets with cognitive distraction annotations. Through extensive evaluations on CogDrive, we show that \name achieves the highest accuracy of 74.38\%, outperforming 11 baselines from 6 model families by over 7\%. Notably, \name can achieve an accuracy of over 70\% across various driving scenarios (different road types, times of day, and weather conditions) with strong generalizability. These results highlight the importance of modeling gaze-context interactions and the effectiveness of cross-modal interaction modeling for multimodal cognitive distraction detection. Our codes and CogDrive dataset resources are available at here\footnote{GitHub repository:\url{https://github.com/langzhang2000/EyeCue}}.

\end{abstract}
\section{Introduction}

\begin{figure}[t]
    \includegraphics[width=0.95\linewidth]{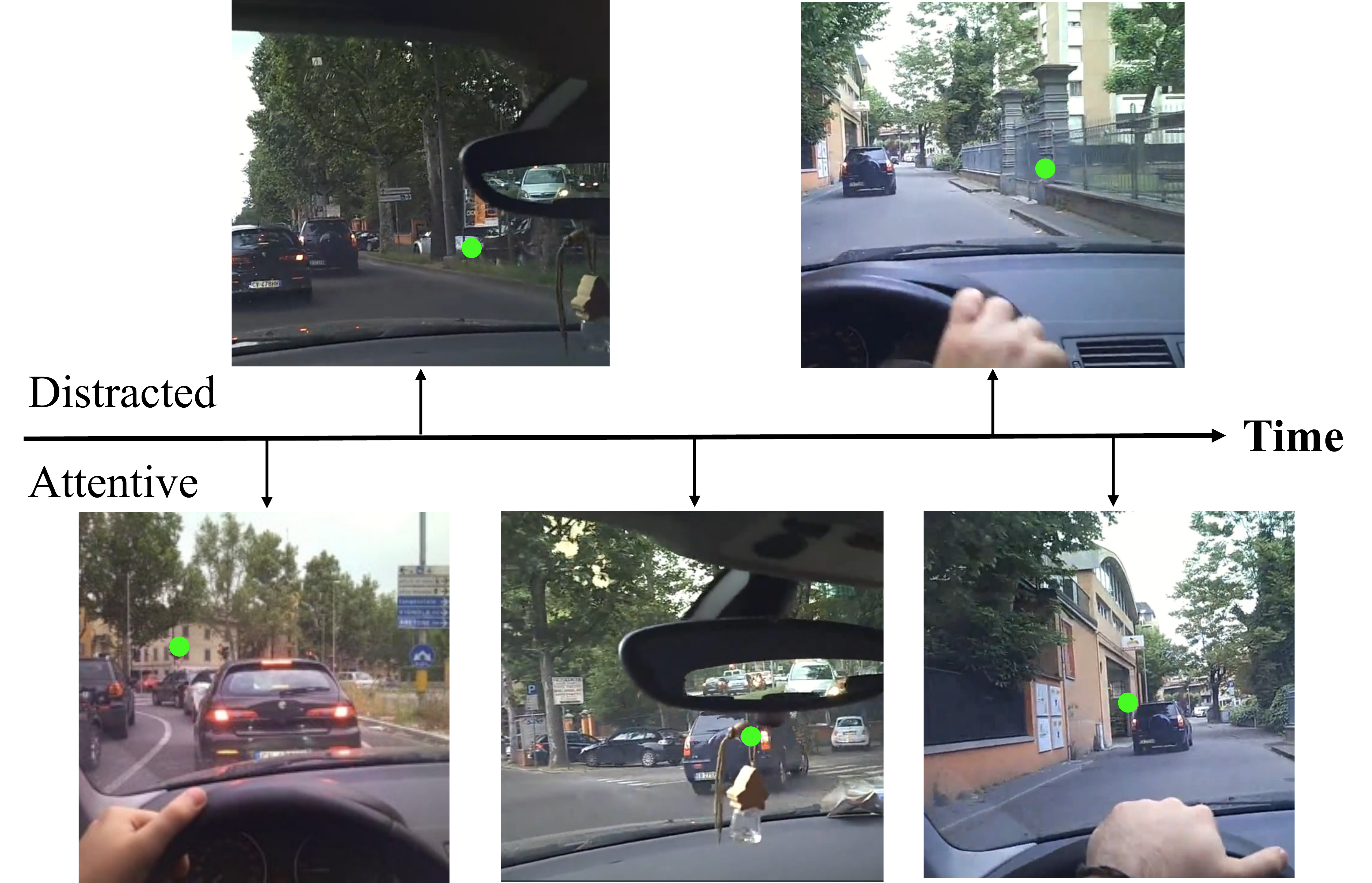}
    \caption{\emph{Is the driver cognitively distracted?} This figure shows a driver's journey on the road, including distracted driving (top) and attentive driving (bottom). Video frames come from the DR(eye)VE dataset~\protect\cite{palazzi2018predicting}, which records the driver’s egocentric views and corresponding gaze points over time. We added a green dot to each raw frame to represent the driver's gaze point at that time.} 
    \label{fig:method_example}
\end{figure}

Distracted driving is a leading cause of traffic fatalities, accounting for approximately 30\% of all such cases, claiming 3,275 lives in 2023~\cite{usdotnhtsa2023fy2024}. It also causes significant property damage, which requires \$9.8 billion in direct infrastructure and road-maintenance costs~\cite{nhtsa2025distracted}. Besides these economic burdens, distracted driving places heavy demands on both law enforcement and medical resources. Therefore, a timely investigation into the problem of driver distraction detection is crucial.

Distractions during driving can be classified into three types: manual, visual, and cognitive distractions~\cite{9405644}: (\romannumeral1) \emph{manual distraction} occurs when the driver's hands are off the wheel (e.g., secondary task: holding a phone or food while driving); (\romannumeral2) \emph{visual distraction} occurs when the driver looks away from the attention areas covered by the windshield or rear‑view mirrors (e.g., looking at a navigation system); (\romannumeral3) \emph{cognitive distraction} occurs when the driver's attention is diverted by thoughts unrelated to driving (e.g., staring at a point while thinking about something else unrelated to driving), even though their gaze remains on the road. Both manual and visual distractions have been widely studied in relation to road safety, whereas cognitive distraction remains underexplored because it is latent and non-observable~\cite{guo2024tla}. We provide a more detailed discussion of these distraction types in Appendix~\ref{type}.

Since cognitive distraction is reflected in the driver's internal mental state over time, it is often difficult to detect directly from external behaviors alone. Even when a driver appears visually attentive (looking at the road) and manually attentive (keeping their hands on the wheel), cognitive distraction may still happen. Existing approaches mostly rely on intrusive methods that require attaching physiological sensors to the driver or interrupting normal driving. For example, electroencephalography (EEG) provides precise neural signals for inferring cognitive state but requires direct physical coupling between sensors and the body~\cite{li2023driver}. Self-reported questionnaires can be used to evaluate a driver’s perception of the driving environment~\cite{cooper2014measuring}, and detection response tasks are commonly used to assess their reaction time to unexpected events~\cite{al2024risk}. However, these methods may disrupt normal driving and only support post-hoc evaluations, limiting their practicality for continuous use. Therefore, \emph{a key research question is how to design a non‐intrusive method to detect driver cognitive distraction over time without disrupting driving.}

To answer this question, we leverage eye gaze as a behavioral cue to understand the driver's cognitive state, as eye gaze reflects how the internal cognitive process allocates attention~\cite{oyama2019novel}. Moreover, with the growing adoption of lightweight augmented reality glasses such as Meta Aria Gen 2, real-time eye-tracking data can be collected in a non-intrusive manner without additional user effort~\cite{engel2023project}. However, relying on eye-tracking data alone for cognitive distraction detection is insufficient, as it lacks contextual scene information~\cite{zhou2025towards}. The same gaze pattern may reflect different cognitive states across driving scenarios. For example, as shown in Fig.~\ref{fig:method_example}, in the first attentive case, the driver fixates on the traffic light while waiting at the signal; in the first distracted case, the driver fixates on a parked vehicle along the roadside during straight‑ahead driving. This naturally motivates a multimodal perspective that adopts both eye gaze and contextual scene information to interpret how attention is temporally allocated during driving.

To that end, we aim to infer the driver’s cognitive state by analyzing temporal eye-gaze patterns and the visual targets of attention using egocentric video, which captures the spatio-temporal structure of the driving scene~\cite{zhou2024embodied}. Specifically, we leverage a key insight: \emph{detecting cognitive distraction requires understanding how the driver’s gaze interacts with the surrounding egocentric visual context over time.} This interaction is reflected both over a short driving clip (i.e., global interaction) and in each frame (i.e., fine-grained interaction). Despite this useful insight, a lack of high-quality datasets suitable for modeling this interaction remains a challenge. Datasets that provide both eye-tracking information and egocentric driving videos with cognitive distraction annotations are very limited, primarily due to the difficulty of data collection and annotation. Hence, we are faced with three key challenges: 
\emph{(C1) How to learn both global and fine-grained features from the driver's egocentric videos and eye gaze data, respectively? 
(C2) How to integrate eye gaze information with egocentric videos to detect driver cognitive distraction? (C3) Lack of scalable and generalizable datasets for cognitive distraction detection.}

To address C1 and C2, we propose \name, a non-intrusive framework that integrates temporal eye gaze with egocentric video to detect driver cognitive distraction. \name consists of three core components:
(\romannumeral1) a \emph{video encoder} that processes each egocentric video clip to understand the surrounding context of the driver; (\romannumeral2) a \emph{gaze encoder} that analyzes eye-tracking data to extract the driver's eye gaze patterns; and (\romannumeral3) a \emph{gaze-driven semantic query (GDSQ)} module that leverages gaze cues to dynamically select visual tokens from the egocentric video, reflecting how the driver’s gaze is allocated across the context over time. Then, we fuse the outputs of these three modules to detect cognitive distraction.
As for C3, DR(eye)VE is the only publicly available dataset that contains eye gaze, egocentric videos, and cognitive distraction labels~\cite{palazzi2018predicting}. However, it covers limited driving scenarios, and the annotated distracted samples may be insufficient. Hence, we explore three other existing driving datasets (along with DR(eye)VE) to create CogDrive, an egocentric cognitive distraction dataset. We select BDD-A~\cite{xia2018predicting}, DADA-2000~\cite{fang2021dada}, and TrafficGaze~\cite{deng2019drivers} since they all provide gaze data and egocentric videos. Following the DR(eye)VE's annotation procedure, we create a larger dataset consisting of 3,662 samples with cognitive distraction annotations.

To the best of our knowledge, \emph{\name is the first work that integrates temporal eye gaze information with egocentric videos to detect driver cognitive distraction.} Our main contributions are summarized as follows:
\begin{itemize}
    \item We propose \name, a gaze-empowered egocentric video framework for cognitive distraction detection that explores the interaction between eye gaze and the driving context. Specifically, \name jointly learns representations from egocentric video and eye gaze through two encoders: a video encoder captures scene context and local visual cues, while a gaze encoder models temporal gaze behavior patterns (addressing Challenge C1). Moreover, we use gaze to guide video preprocessing and introduce a GDSQ module that directs cross-attention toward gaze-relevant visual regions, modeling gaze-context interactions for cognitive understanding (addressing Challenge C2).
    \item We introduce CogDrive, a cognitive distraction dataset consisting of 3,662 annotated egocentric video clips with gaze signals (addressing Challenge C3). It covers various driving scenarios, including diverse road scenes and driving events.
    \item Through extensive experiments on CogDrive, we show that \name achieves an accuracy of 74.38\%, outperforming 11 baselines from 6 model families, including gaze-only, classical video classification, egocentric, foundation, gaze-with-image, and gaze-with-video models, by more than 7\% in absolute gain. These findings highlight the importance of modeling gaze-context interactions and the effectiveness of cross-modal interaction modeling for multimodal cognitive distraction detection.
\end{itemize}
\section{Related Work}

In this section, we first provide the background on driver distraction detection, then discuss gaze-based analysis techniques and video understanding models, and finally, examine recent advances in gaze-empowered vision models.

\paragraph{Driver Distraction Detection.}
There are three main types of driver distraction: manual, visual, and cognitive~\cite{9405644}. For manual and visual distractions, recent approaches commonly use in-car cameras to detect the driver's observable body movements~\cite{sonth2023explainable}. Since cognitive distraction manifests in an individual’s mental state, most research focuses on intrusive solutions to measure their cognitive loads. For example, Figalov{\'a} et al.~\shortcite{figalova2023manipulating} use questionnaires to evaluate the driver's cognitive state. However, these methods could disrupt normal driving. This motivates us to design a non-intrusive method to detect cognitive distractions.

\paragraph{Gaze-based Analysis Techniques.}
The driver's eye gaze cues are valuable for driver attention assessments~\cite{huang2025gaze}. For example, Maralappanavar et al.~\shortcite{maralappanavar2016driver} uses the driver's pupils to estimate the driver's state, but it misses the environmental context. 
Zhou et al.~\shortcite{zhou2025towards} predict the driver's gaze area and generate text explaining why drivers should focus on these areas. Bhagat et al.~\shortcite{bhagat2023driver} show that driver gaze patterns and saliency can vary by driving tasks. Recently, DCDD combines a single frame with eye-tracking data for distraction detection~\cite{qiao2025driver}, but it lacks temporal attention modeling.

\paragraph{Video Understanding Models.}
Video understanding models can understand the context perceived by the driver during driving~\cite{min2024driveworld}. Models like TimeSformer~\cite{bertasius2021space} and VideoMAE~\cite{tong2022videomae} capture the spatio-temporal relationship of visual content. Video foundation models have enhanced reasoning ability when integrated with other modalities (e.g., text)~\cite{wang2024internvideo2}. This inspires us to consider the integration of eye gaze with video understanding models.

\paragraph{Gaze-Empowered Vision Models.}

Current multimodal models illustrate the superiority of integrating eye gaze with visual content. For example, Voila-A~\cite{yan2024voila} and GazeGPT~\cite{konrad2024gazegpt} use eye gaze to help visual models recognize which objects the user is focusing on in an image. GazeLLM reduces the computational load of the model by using only the user's fixation area~\cite{rekimoto2025gazellm}. Egovideo captures the egocentric video for a better perception~\cite{pei2024egovideo}. However, these works focus on the visual content in an image rather than the first-person state understanding in a video. Recently, egoEmotion recognize eye gaze as a critical perceptual modality to reflect human states~\cite{jammot2025egoemotion}. Inspired by this, we design a method that can fuse eye gaze and egocentric video to detect driver cognitive distraction.
\section{Key Insights and Challenges}

In this section, we introduce the key insights and discuss the main challenges in model design and dataset construction.

\subsection{Key Insights}

Research on safe driving shows that the driver’s perception is reflected in how they allocate attention to different objects over time~\cite{du2020psychophysiological}. Musabini and Chetitah~\shortcite{musabini2020heatmap} find that during normal driving, the driver fixates on a wide area of the environment, whereas when cognitively distracted, their attention remains confined to specific regions. In addition, Ojstersek and Topolsek~\shortcite{ojstersek2019eye} observe that the driver looks at different objects in various contexts to maintain an attentive cognitive state. From these studies, we draw the following key insight: \emph{detecting cognitive distraction requires understanding how the driver’s gaze interacts with the surrounding egocentric visual context over time.} When the driver is cognitively attentive, they exhibit more frequent interactions with task-related objects~\cite{bhagat2023driver}. In contrast, when they are distracted, their gaze is less likely to support driving task-related objects in the scene~\cite{sarkar2022comprehensive}.

\subsection{Main Challenges}\label{challenges}

Building on the aforementioned insight, designing a non-intrusive driver cognitive distraction detection system requires addressing two model design challenges and one data construction challenge.

\paragraph{(C1) How to learn both global and fine-grained features from the driver's egocentric view and eye gaze?} 
Specifically, global features help us grasp the overall information about the environment and the driver's gaze throughout the entire video clip. Fine‑grained features reveal object features in the scene and the gaze pattern at each moment. However, global and fine-grained features operate at different temporal and spatial scales. Therefore, we need a way to model these features of gaze and video modalities separately.

\paragraph{(C2) How to integrate gaze information with egocentric videos to detect driver cognitive distraction?}
We need to leverage the driver's eye gaze information to understand how they perceive the environment. For example, the driver might move their gaze from a traffic light to a pedestrian when they are waiting for the red light. In this case, we should fuse gaze information with egocentric video representations to capture how the driver attends to different regions in the scene. How to effectively integrate gaze cues with egocentric video representations remains a major challenge.

\paragraph{(C3) Lack of scalable and generalizable datasets for cognitive distraction detection.} Designing datasets for driver cognitive distraction detection remains challenging due to limited data availability. Currently, DR(eye)VE is the only dataset that provides egocentric driving videos, eye gaze, and cognitive distraction labels. However, DR(eye)VE is collected under restricted driving scenarios and geographic regions, limiting its generalizability to other scenarios. Besides, cognitive distraction is implicit, which makes reliable annotation difficult without a concrete protocol. These challenges highlight both the necessity and difficulty of constructing a comprehensive dataset for cognitive distraction detection.
\section{Our Design: \name}

\begin{figure*}[t]
    \centering
    \includegraphics[width=0.9\linewidth]{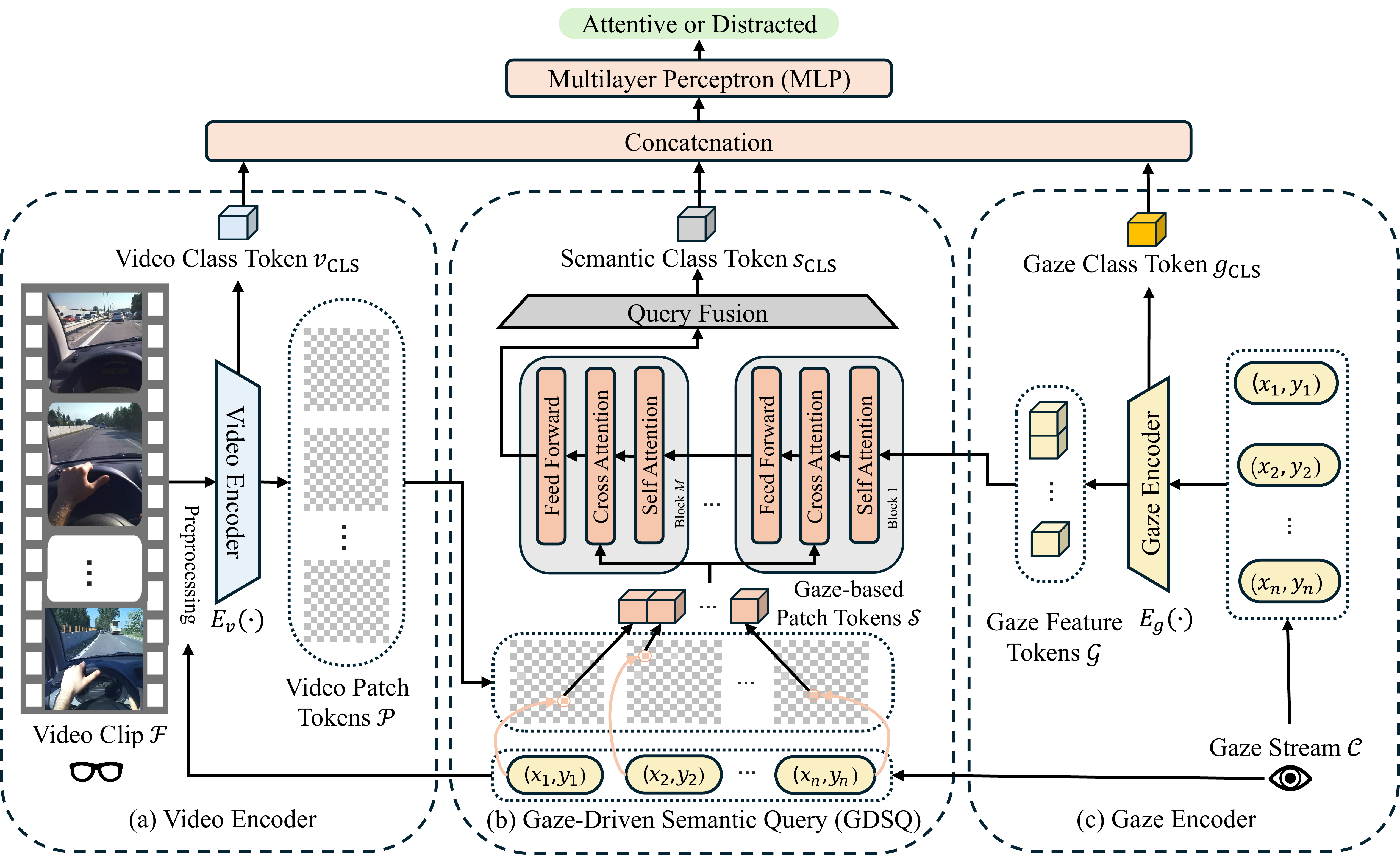}
    \caption{\textbf{\name architecture.} (a) The video encoder extracts the global contextual information and fine-grained visual details. (b) The GDSQ module captures the relationship between visual context and eye gaze. (c) The gaze encoder obtains global attention patterns and frame-level gaze details. Finally, these three types of class tokens are concatenated and fed into a multilayer perceptron for classification.}
    \label{fig:method}
\end{figure*}

In this section, we state our goal and discuss how our solution addresses the aforementioned two design challenges.

\paragraph{Design Goal.}
Our goal is to detect the driver’s cognitive distraction. Based on the driver's egocentric video, we obtain a sequence of preprocessed video clips $\mathcal{F}$ together with temporally aligned gaze coordinates $\mathcal{C}$. The task is formulated as a binary classification problem, where the model predicts whether the driver is \emph{cognitively distracted} or \emph{attentive}.

\paragraph{Architecture Overview.}

We propose \name, a multimodal framework that jointly leverages egocentric video and eye gaze to detect driver cognitive distraction. As illustrated in Fig.~\ref{fig:method}, \name comprises three core components specialized to address the design challenges outlined above.
To address Challenge~C1, \name employs two modality-specific encoders: (\romannumeral1) a \emph{video encoder} that extracts both global and fine-grained spatio-temporal features from the egocentric videos, and (\romannumeral2) a \emph{gaze encoder} that models global attention patterns as well as frame-level gaze dynamics.
These encoders enable the model to represent the driving scene and the driver's gaze behavior across the entire video clip.
To address Challenge~C2, we further introduce (\romannumeral3) a \emph{Gaze-Driven Semantic Query} module, which models the interaction between the gaze and the context. Finally, we concatenate the video class token, gaze class token, and semantic class token to form a unified representation. This representation is then fed into a multilayer perceptron for classification. 

In \name, gaze information is utilized in three ways.
First, in the video encoder, gaze is used to guide video preprocessing by emphasizing regions attended by the driver.
Second, the gaze encoder models both global attention patterns over the entire clip and frame-level gaze dynamics.
Third, the GDSQ module aligns gaze with fine-grained visual content through gaze-guided patch selection and cross-attention.
Together, these designs enable \name to fully exploit both gaze and video modalities for cognitive distraction detection.
Among these strategies, video preprocessing and the GDSQ are critical for modeling the interaction between the driver’s gaze and the surrounding context.

\paragraph{Video Encoder.}

Given an egocentric video clip consisting of $n$ frames, we first apply gaze-based video preprocessing, such as dot overlays, heatmap masks, and cropped frames, to enhance the input videos. Let $f_t$ be the $t$-th frame. Then, we can denote the preprocessed video clip consisting of $n$ frames by $\mathcal{F} := \{f_t\}_{t=1}^n$. Next, we employ a pre-trained video encoder $E_v(\cdot)$ to obtain rich spatio-temporal representations of the driving scene\cite{bertasius2021space}. We prefer a video encoder that can extract features from fine-grained regions within each frame. The video encoder processes the driver's egocentric video to generate a single video class token $v_{\mathrm{CLS}}$ and a sequence of video patch tokens $\mathcal{P} :=\{p_i\}_{i=1}^l$, where $l$ is the total number of video patch tokens, and $p_i$ denotes the $i$-th video patch token. The video class token learns the global representation of the driving environment. Also, each image frame is divided into multiple patches, and each patch is encoded into a patch token to represent the fine‐grained feature of that small area in the environment scene.

\paragraph{Gaze Encoder.}
Let $(x_t,y_t)$ be the aligned eye gaze coordinates in frame $f_t$. Then, for the entire video clip, we have a sequence of coordinates $\mathcal{C} := \{(x_t,y_t)\}_{t=1}^n$. We use a gaze encoder $E_g(\cdot)$ to embed these gaze coordinates into a gaze class token $g_{\mathrm{CLS}}$ and a sequence of gaze feature tokens $\mathcal{G} := \{g_t\}_{t=1}^n$, where $g_t$ denotes the $t$-th gaze feature token. Specifically, in the gaze encoder $E_g(\cdot)$, we first project each fixation coordinate to the same embedding space as the video patch tokens through a learnable linear layer. This produces an embedded coordinates token sequence. Then, we add a class token to this sequence. Next, we process them through a customized lightweight transformer encoder to obtain the gaze class token and the sequence of gaze feature tokens~\cite{vaswani2017attention}. The gaze class token summarizes the driver’s gaze pattern throughout the entire clip. The gaze feature tokens capture the gaze pattern at the frame level.

\paragraph{Gaze-Driven Semantic Query.}

This module integrates the driver's gaze with egocentric videos, which is critical for gaze-context interaction modeling. We aim to extract the driver's visual perception feature. First, we project a gaze point to the corresponding area in the frame and select $h$ video patch tokens around this area. Specifically, for each frame, we select either a single patch (i.e., $h=1$) where the gaze point is located or multiple patches (i.e., $h>1$) that contain the surrounding area. We discuss gaze-based patch token selection strategies in Appendix~\ref{token_selection}. Then, we repeat this operation for all frames, and can get $h\times n$ video patch tokens in total. Gaze-based patch tokens can be denoted by $\mathcal{S}:= \{s_m\}_{m=1}^{h\cdot n}$, where $s_m$ is the $m$-th gaze-based patch token. 

To capture the interaction between gaze and environment at the frame level, we inject gaze feature tokens $\mathcal{G}$ and gaze-based patch tokens $\mathcal{S}$ as input into the cross-attention block. This cross-attention block allows the system to model the interaction between the driver's eye gaze and the environment. We draw on the intuition that the driver uses eye gaze to perceive and query the surrounding environment. Hence, we use gaze feature tokens $\mathcal{G}$ as query inputs, and the gaze-based patch tokens $\mathcal{S}$ serve as key and value inputs in the cross-attention block. We repeat the cross-attention block $M$ times to iteratively refine the relationship between the gaze and the environment features. Then, we apply a pooling strategy to the output of the $M$-th cross-attention block to obtain a single semantic class token $s_{\mathrm{CLS}}$. The class token can represent the overall interaction between the driver and the context. 

Finally, we concatenate the three types of class tokens ($v_{\mathrm{CLS}},\ s_{\mathrm{CLS}}$, and $g_{\mathrm{CLS}}$), which capture global information for classification, and feed them into a multilayer perceptron that predicts whether the driver is cognitively distracted.

\section{Evaluations}
\label{sec:evaluations}

In this section, we describe the implementation details and experimental setup, introduce the CogDrive dataset, and present evaluation results for \name.

\subsection{Implementation and Experimental Setup}

\paragraph{Model Configuration.}
For the \emph{video encoder}, we evaluate two pre-trained backbones: TimeSformer$_{\text{K600}}$~\cite{bertasius2021space} and VideoMAE$_{\text{K400}}$~\cite{tong2022videomae}. The \emph{gaze encoder} consists of a Transformer with 8 attention heads and 1 encoder block to embed raw gaze information. The \emph{GDSQ} module has 2 cross-attention blocks (i.e., $M=2$), followed by a two-layer MLP. The hyperparameters of the model configuration are chosen empirically, as this lightweight setting achieves strong performance. We discuss video encoder selection strategies in Appendix~\ref{video_selection}.

\paragraph{Baselines.}
We compare \name with 11 baselines from 6 model families (i) \emph{Gaze only}: Heatmap-based SVM~\cite{musabini2020heatmap}; (ii) \emph{Classical backbones}: TimeSformer~\cite{bertasius2021space} and  VideoMAE~\cite{tong2022videomae}; (iii) \emph{Egocentric video understanding models}: EgoVideo~\cite{pei2024egovideo}; (iv) \emph{Video foundation models}: InternVideo2~\cite{wang2024internvideo2}, Video-LLaVA~\cite{lin2023video}, and VideoLLaMA3~\cite{zhang2025videollama}; (v) \emph{Gaze with images}: GazeGPT~\cite{konrad2024gazegpt} and Voila-A~\cite{yan2024voila}; (vi) \emph{Gaze with videos}: GazeVQA~\cite{ilaslan2023gazevqa} and GazeLLM~\cite{rekimoto2025gazellm}.

\paragraph{Training and Inference.}
We conduct training and inference on an NVIDIA L40S GPU, fine-tuning the video encoder, and training the gaze encoder and GDSQ module from scratch for 15 epochs. The remaining training configuration settings follow the default specifications of the video encoder. All trainable baseline methods are fine-tuned on the CogDrive dataset for fair comparison, while training-free methods (e.g., GazeGPT and GazeLLM) are evaluated without fine-tuning. We discuss prompt designs for language models in Appendix~\ref{prompt}. Trainable baselines are averaged over five runs, while training-free models are evaluated once.

\subsection{CogDrive Dataset}
\begin{table}[t]
\centering
\begin{tabular}{lrrr}
\toprule
\textbf{Dataset} & \textbf{Attentive} & \textbf{Distracted} & \textbf{Total} \\
\midrule
DR(eye)VE   & 1,485 (78.41\%) & 409 (21.59\%) & \textbf{1,894}   \\
BDD-A       & 424 (66.88\%) & 210 (33.12\%) & \textbf{634}  \\
DADA-2000   & 463 (73.84\%) & 164 (26.16\%) & \textbf{627}  \\
TrafficGaze & 481 (94.87\%) & 26 (5.13\%) & \textbf{507}   \\
\midrule
\textbf{CogDrive} & \textbf{2,853 (77.91\%)} & \textbf{809 (22.09\%)} & \textbf{3,662} \\
\bottomrule
\end{tabular}
\caption{Clip statistics of CogDrive datasets.}
\label{tab:cogdrive_stats}
\end{table}
To address Challenge C3, we introduce CogDrive, a comprehensive dataset built by integrating four existing driving video datasets: DR(eye)VE~\cite{palazzi2018predicting}, BDD-A~\cite{xia2018predicting}, DADA-2000~\cite{fang2021dada}, and TrafficGaze~\cite{deng2019drivers}. We select these datasets because they provide both gaze and egocentric videos. We follow a unified protocol derived from DR(eye)VE's labels with domain expert guidance for consistency. Detailed annotation procedures are provided in Appendix~\ref{annotation}. The protocol defines attentive vs. distracted criteria based on gaze-context interaction. All clips are independently labeled by two annotators and subsequently checked by a domain expert, achieving an inter-annotator agreement of over $98\%$. Following prior observations that cognitive distraction clips are typically short, we segment videos into fixed-length clips of $16$ frames.

\begin{table*}[t]
  \centering
  \begin{tabular}{lllcccccccc}
  \toprule
    \multicolumn{2}{c}{\multirow{4}{*}{\textbf{Methods}}} & \multirow{4}{*}{\textbf{Backbone}} & \multicolumn{6}{c}{\textbf{Leave-one-dataset-out}} & \multicolumn{2}{c}{\textbf{Aggregated}} \\
    \cmidrule(lr){4-9}\cmidrule(lr){10-11}& & &\multicolumn{2}{c}{\textbf{BDD-A}} & \multicolumn{2}{c}{\textbf{DADA-2000}} & \multicolumn{2}{c}{\textbf{DR(eye)VE}} & \multicolumn{2}{c}{\textbf{CogDrive}} \\\cmidrule(lr){4-5}\cmidrule(lr){6-7}\cmidrule(lr){8-9}\cmidrule(lr){10-11}
     & & & \textbf{Acc.} & \textbf{F1} & \textbf{Acc.} & \textbf{F1} & \textbf{Acc.} & \textbf{F1} & \textbf{Acc.} & \textbf{F1}
    \\
    \midrule
    Gaze-only
     & Heatmap-based & SVM & \underline{61.95} & 0.57 & 65.85 & 0.64 & 49.39 & 0.14 & 66.12 & 0.64 \\ 
    \midrule
    \multirow{3}{*}{Classical}
      & $\text{TimeSformer}_{\text{K400}}$  &  ViT-B/16 & 60.71 & 0.62 & 62.63 & 0.63 & 55.40 & 0.55 & 65.35 & 0.70 \\ 
      & $\text{TimeSformer}_{\text{K600}}$  &  ViT-B/16 & 60.52 & \underline{0.67} & 61.89 & 0.53 & 55.41 & 0.61 & 66.80 & \underline{0.71} \\ 
      & $\text{VideoMAE}_{\text{K400}}$     &  ViT-B/16 & 57.38 & 0.62 & 61.28 & 0.60 & 58.85 & 0.64 & 67.21 & 0.66 \\ 
    \midrule
    Egocentric
      & EgoVideo &  ViT-B/14 & 58.57 & 0.65 & 59.15 & 0.63 & 55.65 & 0.56 & 65.29 & 0.68 \\ 
    \midrule
    \multirow{3}{*}{Foundation}
    & $\text{InternVideo2}_\text{s1-1B}$ &  ViT-B/14 & 50.16 & 0.45 & 50.51 & 0.45 & \underline{59.31} & 0.58 & 56.61 & 0.53 \\
      & Video-LLaVA  & OpenCLIP-L/14 & 52.43 & 0.37 & 52.09 & 0.47 & 53.75 & 0.51 & 55.06 & 0.55  \\ 
      & VideoLLaMA3   & ViT-B/16 & 54.86  & 0.47  & 53.26 & 0.38 & 51.17 & 0.39 & 53.17 & 0.45 \\
    \midrule
    \multirow{2}{*}{Gaze w/ images}
     & GazeGPT$^*$  & GLM-4.5V-AWQ  & 51.00 & 0.47 & 51.81 & 0.53 & 51.29 & 0.42 & 51.69 & 0.48  \\ 
     & Voila-A   & CLIP ViT-L/14 & 58.81 & 0.57 & \underline{67.68} & \underline{0.67} & 54.91 & 0.45 & 62.81 & 0.58\\
    \midrule
    \multirow{4}{*}{Gaze w/ videos}
    & GazeVQA & CLIP ViT-B/32 & 59.15 & 0.49  & 59.45 & 0.47 & 51.97 & \underline{0.65} & 67.77 & 0.68 \\
    & GazeLLM$^*$ & Gemini-2.5-Pro & 44.53 & 0.02 & 49.09 & 0.28 & 49.14 & 0.08 & 48.35 & 0.12 \\
    & $\textbf{\name (Ours)}$ & $\text{VideoMAE}_{\text{K400}}$ & 60.95 & 0.59 & 62.50 &  0.62 & 54.67 & 0.62 & \underline{70.83} & 0.68 \\
    & $\textbf{\name (Ours)}$ & $\text{TimeSformer}_{\text{K600}}$ & \textbf{65.24} & \textbf{0.71} & \textbf{68.29} & \textbf{0.68} & \textbf{60.20} & \textbf{0.67}  & \textbf{74.38} & \textbf{0.74} \\
    \bottomrule
  \end{tabular}
    \caption{Comparison of accuracy (\%) and F1 score between \name and baseline methods. Methods marked with $*$ are training-free models. \textbf{Bold} and \underline{underlined} values indicate the best and the second-best results, respectively.}
  \label{tab:comparison}
\end{table*}

\paragraph{Clip-Level Statistics.}
Table~\ref{tab:cogdrive_stats} reports the clip statistics of CogDrive. The dataset contains 3,662 clips, including 2,853 attentive and 809 distracted samples. Attentive clips dominate across all source datasets, which is in line with reality. For training, we use all 809 distracted clips and randomly select 809 attentive clips following the sub-dataset distributions, then split the data into 70\% for training and 30\% for testing.

\begin{figure}[t]
    \centering
    \includegraphics[width=0.48\textwidth]{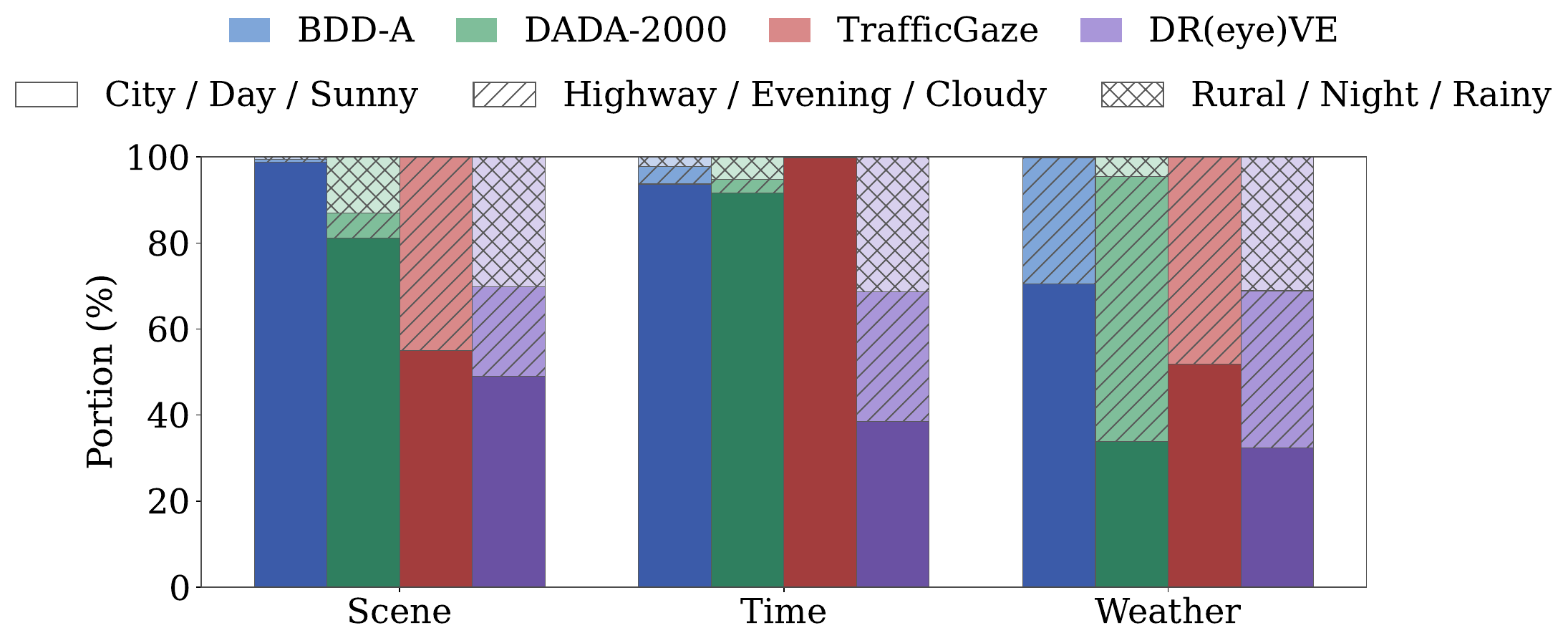}
    \caption{Clip distribution of CogDrive across various scenarios.}
    \label{fig:cogdrive_scene}
\end{figure}

\paragraph{Scenario-Wise Distribution.}
Fig.~\ref{fig:cogdrive_scene} illustrates the clip distribution of CogDrive across four datasets under three complementary conditions: road type, time of day, and weather. This visualization highlights both intra-dataset composition and cross-dataset heterogeneity.

\subsection{Evaluation Results}
\paragraph{Comparison Study.}

Table~\ref{tab:comparison} reports the accuracy and F1 score (for the distracted class) of \name and 11 competitive baselines from 6 model families evaluated on the full CogDrive dataset and its sub-datasets. We adopt a leave-one-dataset-out protocol to assess cross-dataset generalization (e.g., training on DADA-2000, DR(eye)VE, and TrafficGaze when evaluating on BDD-A). We do not do leave-one-dataset-out on TrafficGaze due to its severe class imbalance. Overall, \name achieves the best performance across all settings. On the full CogDrive dataset, \name with a pre-trained TimeSformer$_{\text{K600}}$ backbone achieves the highest accuracy of 74.38\% and an F1 score of 0.74, outperforming all baselines; it also demonstrates strong generalization under the leave-one-dataset-out setting. Notably, models trained with DR(eye)VE included in the training set generalize better than those trained on BDD-A or DADA-2000, suggesting that DR(eye)VE provides more informative samples for cognitive distraction modeling. In contrast, gaze-only, classical video methods, egocentric video models, video foundation models, and gaze-integrated vision approaches all show worse performance, indicating that spatio-temporal visual modeling or gaze cues alone are insufficient. These results highlight the importance of modeling gaze-context interactions and demonstrate the strong generalizability of \name.

\paragraph{Gaze-Guided Video Preprocessing.}\label{gaze_augmentation}

\begin{figure}[t]           
  \centering
  \setlength{\tabcolsep}{5pt} 
  \renewcommand{\arraystretch}{1.0} 
  \begin{tabular}{ccc}       
    \begin{subfigure}{0.14\textwidth}
      \centering
      \includegraphics[width=\linewidth]{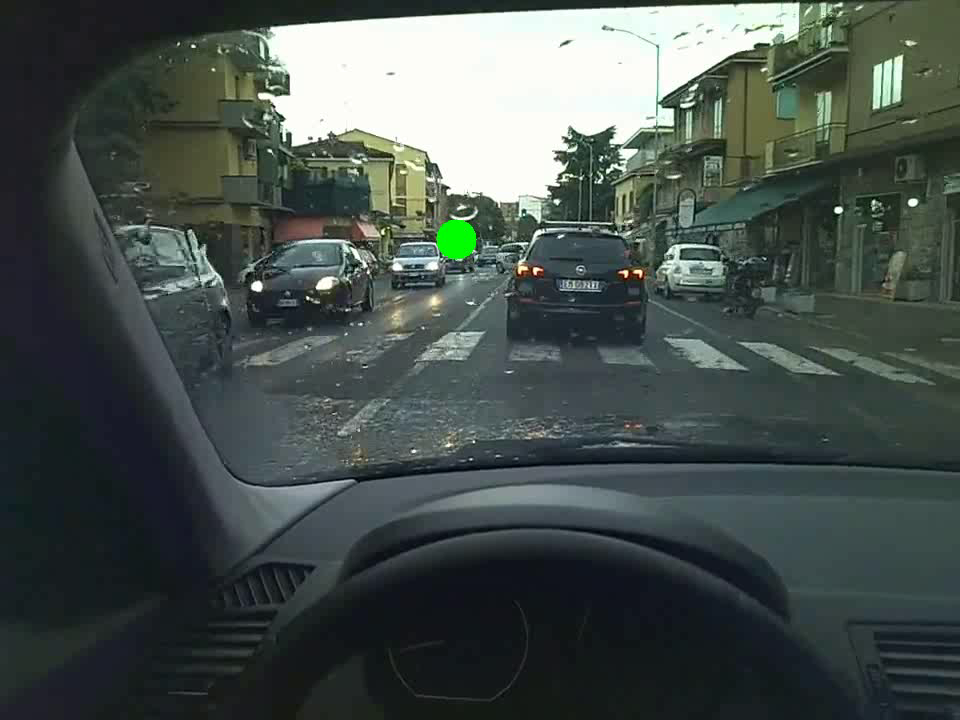}
      \caption{Dot overlay \quad (20 pixels)}
      \label{fig:aug_b}
    \end{subfigure} &
    \begin{subfigure}{0.14\textwidth}
      \centering
      \includegraphics[width=\linewidth]{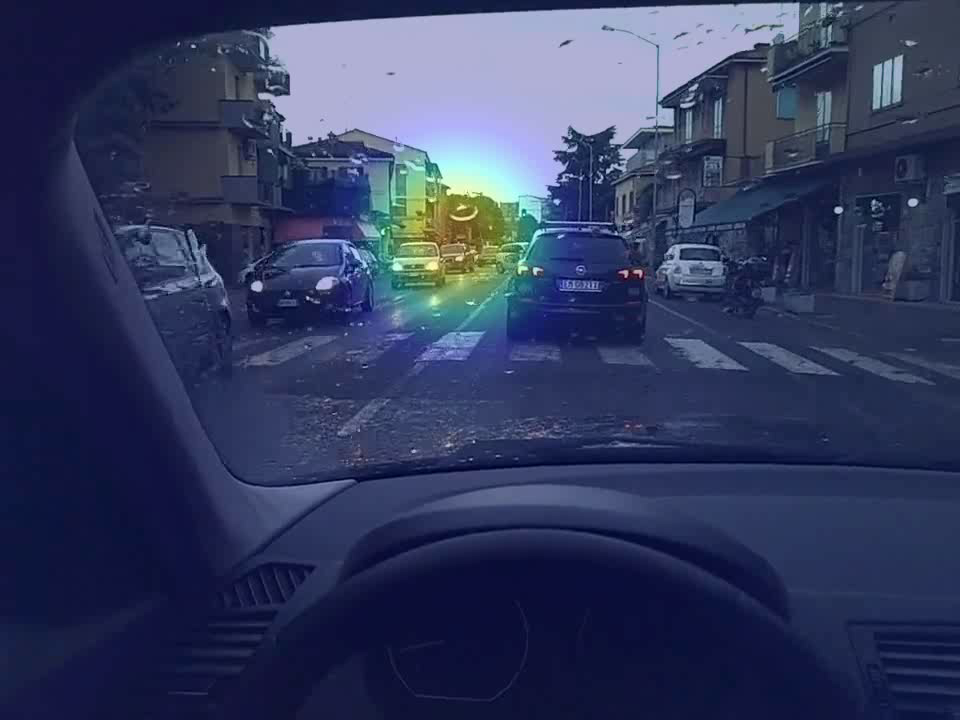}
      \caption{Heatmap mask (75 pixels)}
      \label{fig:aug_d}
    \end{subfigure} &
    \begin{subfigure}{0.13\textwidth}
      \centering
      \includegraphics[width=\linewidth]{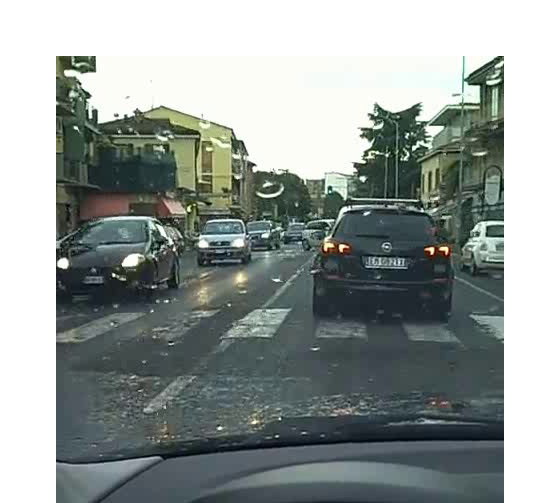}
      \caption{Cropped frame (448$\times$448 pixels)}
      \label{fig:aug_f}
    \end{subfigure}
  \end{tabular}
  \caption{Example of gaze-integrated video preprocessing methods.}
  \label{fig:gaze_aug}
\end{figure}

\begin{figure}[t]
    \centering
    \includegraphics[width=0.99\linewidth]{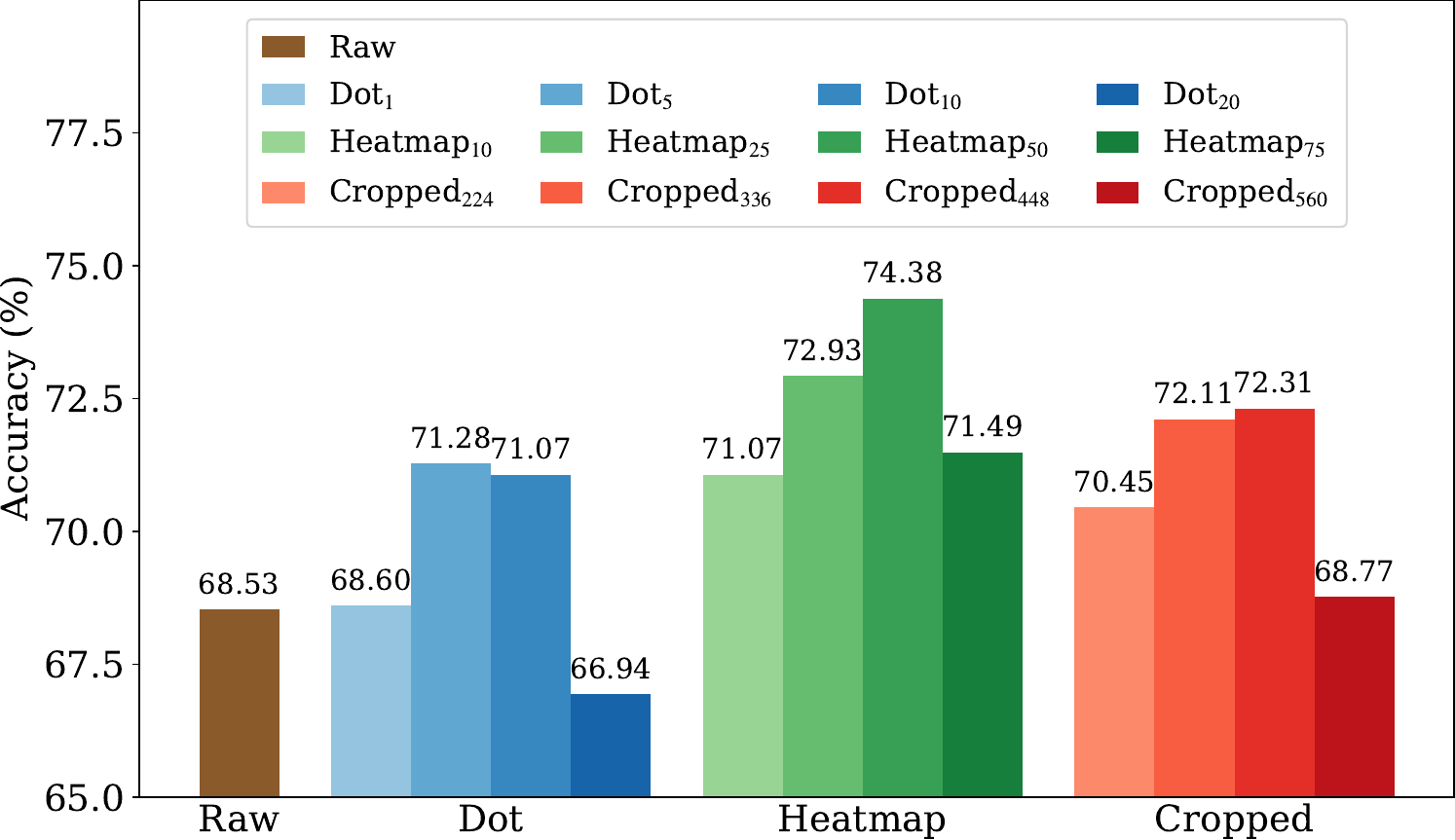}
    \caption{Accuracy (\%) under different preprocessing methods.}
    \label{fig:aug_acc}
\end{figure}

Beyond raw videos, we explore several gaze-based video preprocessing strategies (Fig.~\ref{fig:gaze_aug}), including dot overlays, heatmap masks, and gaze-centered cropping. As shown in Fig.~\ref{fig:aug_acc}, heatmap masks perform best, achieving the highest accuracy with a medium kernel radius, as they emphasize gaze-attended regions while preserving surrounding context. Dot overlays offer limited gains and degrade accuracy when overly salient. Cropping improves accuracy only within a narrow range of crop sizes, reflecting a trade-off between local focus and global context.

\paragraph{Hyperparameter Analysis.}\label{select}

We analyze two key hyperparameters in \name: clip length and the number of gaze-based video tokens. We compare clip lengths of 8 and 16 frames, and vary the number of gaze-based tokens as $h\in\{1,5,9,25\}$ to control regional coverage around each fixation. As shown in Table~\ref{tab:sensitivity}, \name achieves the highest accuracy at 74.38\% with 16-frame clips and a single gaze-based token per frame (i.e., $h=1$). Overall, 16-frame inputs get higher accuracy results, indicating the benefit of longer temporal context. In contrast, increasing $h$ leads to a gradual accuracy drop, reaching its lowest value at $h=25$. We attribute this phenomenon to the foveated nature of human vision. When $h = 1$, the fixation-centered token corresponds to the single video patch that contains the gaze point in each frame, providing a precise representation of the driver's foveal visual focus. In contrast, larger $h$ selects multiple surrounding patches, expanding the representation beyond the fovea and introducing spatial overlap across frames, which dilutes the eye cue and weakens gaze-context interactions.

\begin{table}[t]
\centering
\small
\setlength{\tabcolsep}{6pt}
\renewcommand{\arraystretch}{1.05}
\setlength{\tabcolsep}{10pt}
\begin{tabular}{ccccc}
\toprule
\multirow{2.5}{*}{\textbf{\# of Frames}} &
\multicolumn{4}{c}{\textbf{Gaze-based Video Tokens} } \\
\cmidrule(lr){2-5}
& $h=1$ & $h=5$ & $h=9$ & $h=25$ \\
\midrule
8  & 74.09 & 72.93 & 70.04 & 68.60 \\
16 & \textbf{74.38} & 73.35 & 72.11 & 70.04 \\
\bottomrule
\end{tabular}
\caption{Accuracy (\%) under different number of input frames and gaze-based video tokens.}
\label{tab:sensitivity}
\end{table}

\paragraph{Ablation Study.}

\begin{table}[t]
\centering
\small
\setlength{\tabcolsep}{4.5pt}
\begin{tabular}{l|ccccccc}
\toprule
\textbf{Gaze}  & $\checkmark$ &  &  & $\checkmark$ &  & $\checkmark$ & $\checkmark$ \\
\textbf{Video} &  & $\checkmark$ &  &  & $\checkmark$ & $\checkmark$ & $\checkmark$ \\
\textbf{GDSQ}  &  &  & $\checkmark$ & $\checkmark$ & $\checkmark$ &  & $\checkmark$ \\
\midrule
\textbf{Acc.}  & 54.13 & 67.53 & 68.80 & 69.36 & 70.25 & 72.31 & \textbf{74.38} \\
\bottomrule
\end{tabular}
\caption{Accuracy (\%) of ablation study for \name components.}
\label{tab:ablation}
\end{table}

We conduct an ablation study to analyze the contribution of each component in \name. The results are shown in Table~\ref{tab:ablation}. Using only the gaze encoder yields limited performance as 54.13\%, while the video encoder alone achieves 67.53\%, highlighting the importance of scene context. Using the GDSQ module alone improves accuracy to 68.80\%, suggesting that explicitly modeling gaze–context interactions captures useful information for cognitive distraction recognition.
Direct fusion of the video and gaze encoders further improves accuracy to 72.31\%, indicating that jointly modeling the two modalities provides complementary cues. Hence, both GDSQ and direct fusion highlight the importance of incorporating video and gaze information. Finally, the full \name model, which combines direct multimodal fusion with gaze-context interaction modeling through GDSQ, achieves the best accuracy of 74.38\%. In all settings with GDSQ, both video and gaze encoders are executed, but their class tokens may not be used. Overall, these results highlight the importance of interactions between gaze and context.

\paragraph{Scenario Analysis.}

\begin{table}[t]
\centering
\small
\setlength{\tabcolsep}{2.1pt}
\begin{tabular}{l c c c c c c c c c}
\toprule
& \multicolumn{3}{c}{\textbf{Scene}} & \multicolumn{3}{c}{\textbf{Time}} & \multicolumn{3}{c}{\textbf{Weather}} \\
\cmidrule(lr){2-4}\cmidrule(lr){5-7}\cmidrule(lr){8-10}
& C & H & R & D & E & N & Su & Cl & Ra \\
\midrule
\textbf{Clips} & 368 & 50 & 66 & 366 & 44 & 74 & 253 & 147 & 84 \\
\textbf{Acc.} & 73.64 & 76.00 & 78.79 & 74.32 & 61.36 & 79.73 & 74.70 & 72.79 & 72.62 \\
\bottomrule
\end{tabular}
\caption{Scenario-wise accuracy (\%). “Clips” denotes the number of samples. C=City, H=Highway, R=Rural; D=Day, E=Evening, N=Night; Su=Sunny, Cl=Cloudy, Ra=Rainy.}
\label{tab:breakdown_compact}
\end{table}

We report the scenario-wise accuracy of \name in Table~\ref{tab:breakdown_compact} across different driving scenes, times of day, and weather conditions. \name performs more reliably in rural scenarios. This may be because city driving involves more distracting factors, while the monotonous highway environment makes the driver's mind wander. Regarding time of day, night driving yields higher accuracy, likely because the scene generally includes fewer visible objects. As for weather, sunny scenes yield the highest accuracy due to its better visual quality. Overall, these results suggest that the model generalizes well across diverse real-world scenarios. We provide detailed failure cases analysis in Appendix~\ref{scenario_analysis}.

\paragraph{Additional Evaluation Results.}
We provide confusion matrices, Receiver Operating Characteristic (ROC), Area Under the Curve (AUC), computational complexity, and gaze noise robustness analysis for \name in Appendix~\ref{more_evaluation}.
\section{Conclusion}
We propose \name, a gaze-empowered egocentric video understanding model, to detect the driver's cognitive distraction in a non-intrusive way. By explicitly integrating eye gaze with egocentric video, \name models the interaction between the driver and the surrounding context. Extensive experiments show that \name achieves strong performance. Besides, the CogDrive dataset establishes a new benchmark for cognitive distraction across diverse driving scenarios.

Despite these advantages, \name has several limitations. First, the additional datasets used to construct CogDrive are not originally designed for cognitive distraction analysis, which may lead to less consistent cognitive cues. Second, \name may fail in complex driving scenes and poor gaze data quality situation. Addressing these limitations will further improve robustness in real-world driving scenarios.

\section*{Ethical Statements}
All data used in this work is obtained from publicly available datasets. We do not collect any new human-subject data. In addition, our method only uses gaze coordinates as input and does not involve biometric information such as iris patterns, pupil images, or other personally identifiable ocular features.

\section*{Acknowledgments}
We thank the Virginia Tech Transportation Institute (VTTI) for their support with the dataset annotation. This work was supported by Inha University Research Grant.

\bibliographystyle{named}
\bibliography{ijcai26}

\clearpage
\newpage

\appendix
\section{Appendix}

\subsection{Types of Driver Distraction}\label{type}

There are three types of driver distraction~\cite{9405644}: (i) manual; (ii) visual; and (iii) cognitive. 

\begin{figure}[t]
  \centering
  \setlength{\tabcolsep}{5pt} 
  \renewcommand{\arraystretch}{1.0} 
  \begin{tabular}{ccc}       
    \begin{subfigure}{0.13\textwidth}
      \centering
      \includegraphics[width=\linewidth]{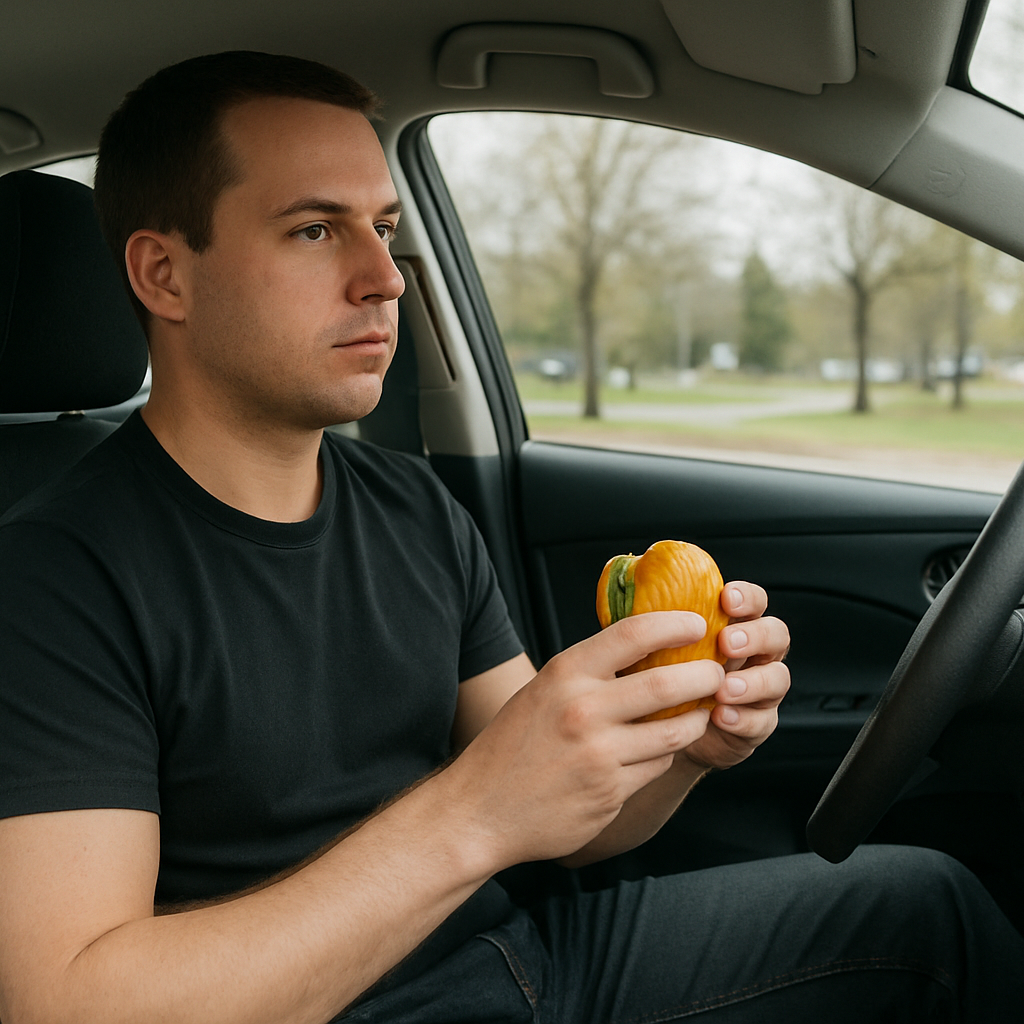}
      \caption{Manual}
      \label{fig:manual}
    \end{subfigure} &
    \begin{subfigure}{0.13\textwidth}
      \centering
      \includegraphics[width=\linewidth]{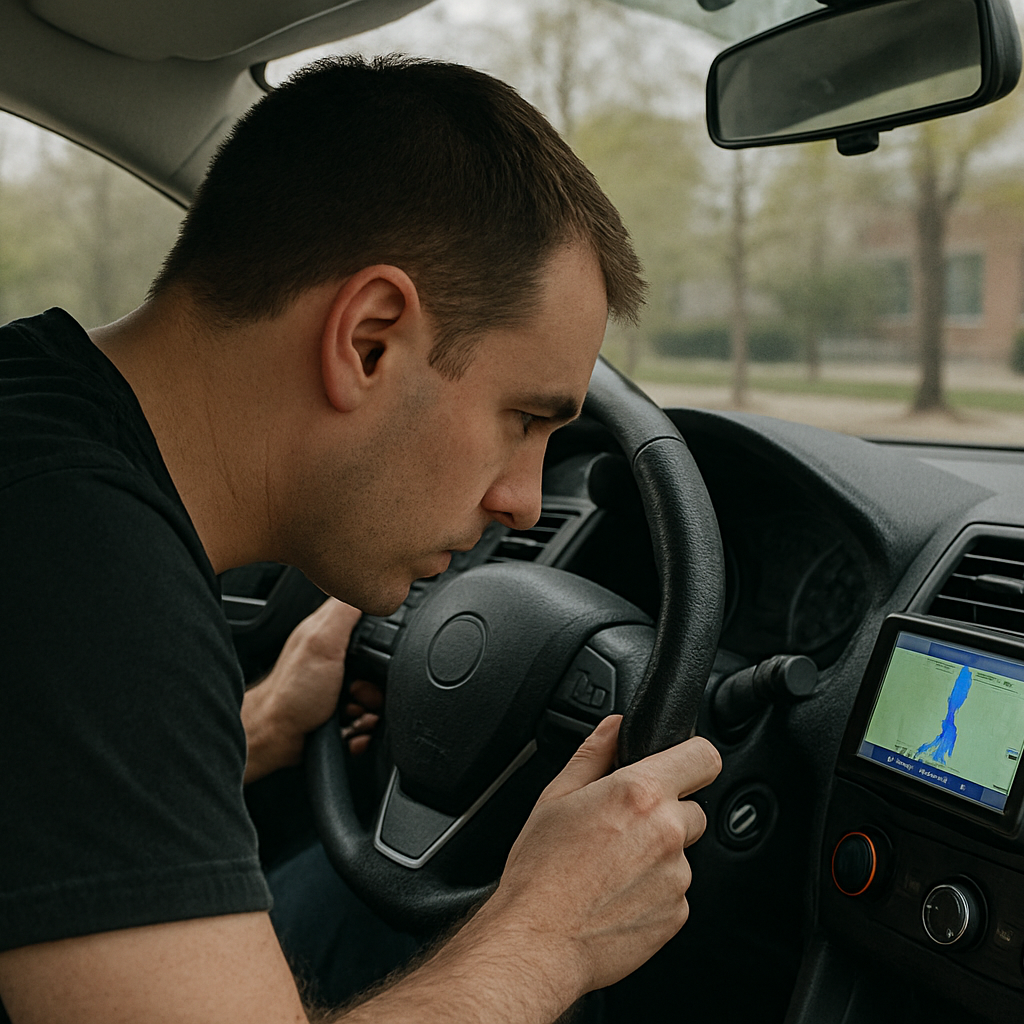}
      \caption{Visual}
      \label{fig:visual}
    \end{subfigure} &
    \begin{subfigure}{0.13\textwidth}
      \centering
      \includegraphics[width=\linewidth]{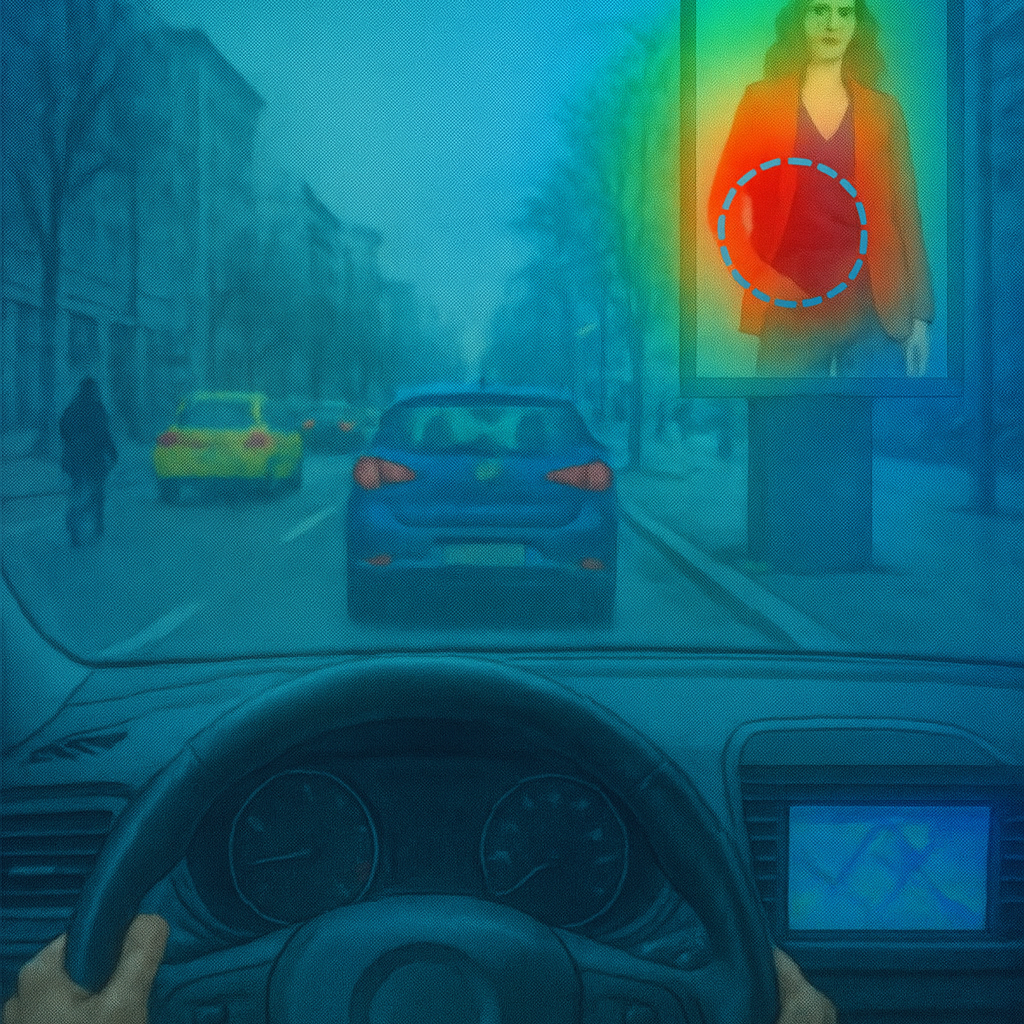}
      \caption{Cognitive}
      \label{fig:cognitive}
    \end{subfigure}
  \end{tabular}
  \caption{Examples of three types of driver distraction. These figures are created using OpenAI's ChatGPT (GPT-4o) with image generation powered by DALL·E 3.}
  \label{fig:disraction}
\end{figure}

\paragraph{Manual Distraction.} Manual distraction occurs when the driver temporarily removes one or both hands from the steering wheel during vehicle operation to perform tasks unrelated to driving (e.g., adjusting the in‑vehicle multimedia system, operating navigation devices, or handling personal items). This condition compromises the driver’s ability to maintain precise control of the vehicle. Fig.~(\ref{fig:manual}) depicts a manual distraction case, where the driver has removed both hands from the steering wheel to hold the food.

\paragraph{Visual Distraction.} Visual distraction occurs in any situation when the driver's line of sight deviates from the attention area required for the safe operation of the vehicle. Specifically, their attention is away from the windshield, rear, or side view mirror area (such as looking at the navigation screen, checking a smartphone, or focusing on the view through a side window). This deviation of visual attention from driving-relevant areas weakens the perception of the driving scene. Fig.~(\ref{fig:visual}) depicts a visual distraction case, where the driver is looking at the navigation screen. 

\paragraph{Cognitive Distraction.} Cognitive distraction is when the driver thinks about things unrelated to driving (e.g., having conversations or daydreaming). It refers to a state in which the driver's gaze remains within the windshield, rear, or side view mirror area; however, their gaze is not directed to the objects relevant to the driving task. In this condition, the driver is unable to allocate sufficient cognitive attention to perceive and interpret the overall driving environment. In Fig.~(\ref{fig:cognitive}), we overlay an eye-based heatmap and depict a case of cognitive distraction, in which the driver focuses on the advertisement while ignoring the vehicle ahead.

Note that the three types of driver distraction may occur concurrently. Here, we mainly focus on cognitive distraction.

\subsection{Implementation Details}

\paragraph{Gaze-based Patch Token Selection.}\label{token_selection}
\begin{figure}[t]
    \centering
    \includegraphics[width=0.99\linewidth]{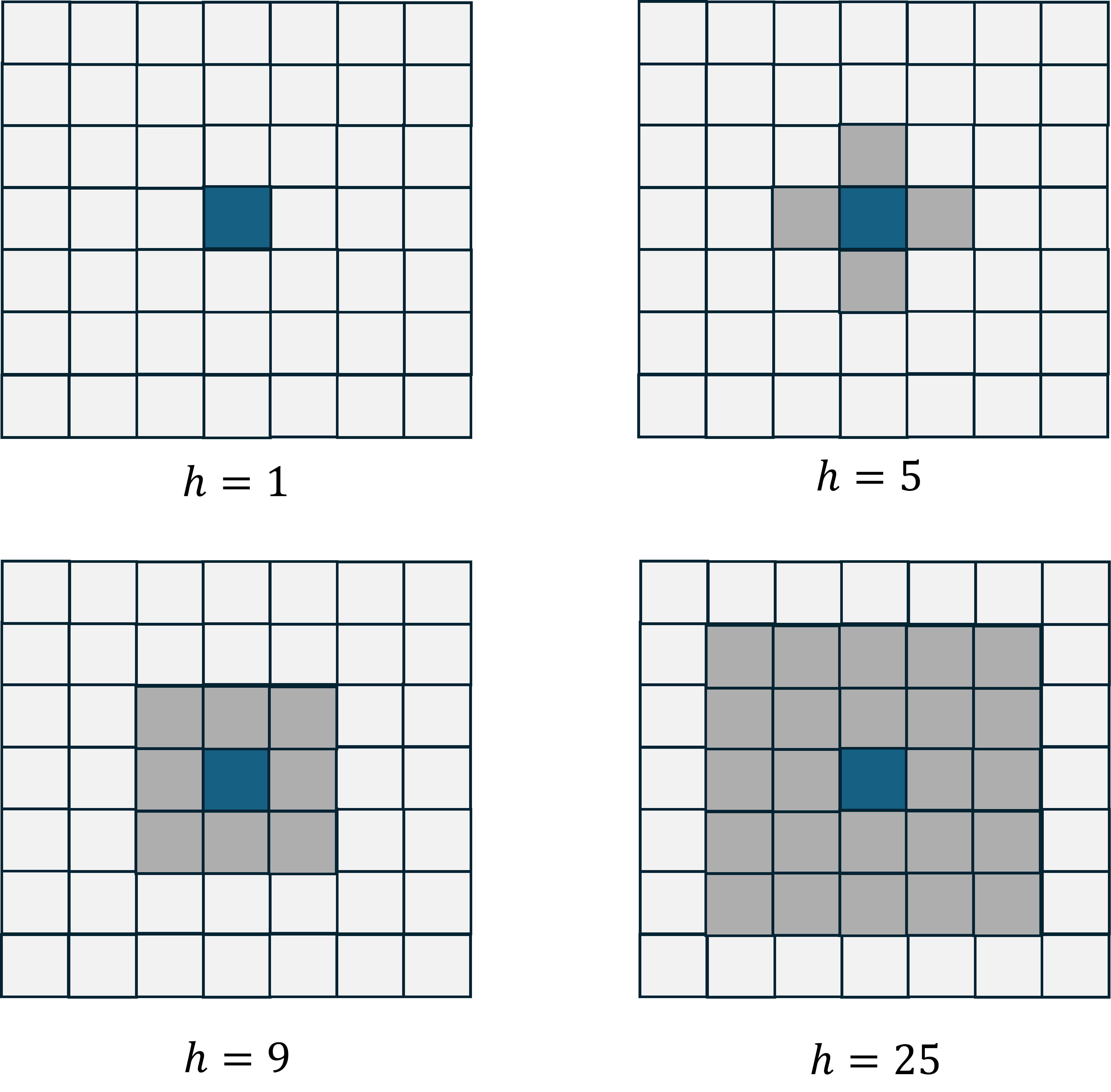}
    \caption{Examples of selected gaze‑based patch tokens when the driver is gazing at the center of the frame. The blue square is where the driver's gaze point belongs, while the gray squares are contextual patches around the gaze point.} 
    \label{fig:patch}
\end{figure}

\begin{table*}[!t]
\centering
\small
\begin{tabular}{llll}
\toprule
\textbf{Scene} & \textbf{Driving Behavior} & \textbf{Gaze Behavior} & \textbf{Label} \\
\midrule
\multirow{4}{*}{City}
 & 
 & Far road center, vehicle ahead, road sign 
 & Attentive \\
 \cmidrule{3-4}&  Driving straight
 & Roadside advertisement, roadside building
 & Distracted \\
\cmidrule{3-4}\multicolumn{4}{c}{\dots} \\
\midrule
\multirow{4}{*}{Highway}
 & 
 & Far road center, lead vehicle 
 & Attentive \\
 \cmidrule{3-4}&  Driving straight
 & Roadside barrier, roadside environment 
 & Distracted \\
\cmidrule{3-4}\multicolumn{4}{c}{\dots} \\
\midrule
\multirow{4}{*}{Rural}
 & 
 & Forward road, roadside cues 
 & Attentive \\
 \cmidrule{3-4}& Turning at the intersection
 & Roadside vegetation, irrelevant objects 
 & Distracted \\
\cmidrule{3-4}\multicolumn{4}{c}{\dots} \\
\bottomrule
\end{tabular}
\caption{Cognitive attention annotation protocol used to assist expert labeling.
This table summarizes representative combinations of driving behaviors and gaze focus areas under different scenes.
We provide a more comprehensive protocol, covering additional driving behaviors and gaze patterns, in a separate file, named ``cognitive attention protocol.xlsx''.}
\label{tab:attention_protocol}
\end{table*}

We apply gaze-based patch token selection in the GDSQ module to select a subset of patch tokens from the output of the video encoder.
Fig.~\ref{fig:patch} illustrates how we use eye gaze information to select patch tokens from the video patch tokens $\mathcal{P}$ in a single frame. When $h=1$, we only select the patch at the driver's fixation point; when $h=5$, we select the fixation patch plus its four immediate neighbors; when $h=9$, we select the \(3\times3\) patches around the gaze point; when $h = 25$, we select the \(5\times5\) patches around the gaze point.

\paragraph{Video Encoder Backbone Selection.}\label{video_selection}
A suitable video encoder is crucial for context capturing from the driver's fixation regions. We prefer backbones that can generate dense patch tokens in each frame, such as TimeSformer~\cite{bertasius2021space} and VideoMAE~\cite{tong2022videomae}, as finer-grained tokens allow more precise localization of the driver’s gaze region and better modeling of gaze-context interaction. In contrast, some video encoders employ token reduction or rely solely on a few summary tokens to boost efficiency~\cite{arif2025hired,hao2025principles}. This limits fine-grained gaze representation, which might fail to capture the detailed interaction between the driver's eye gaze and the driving scene.

\paragraph{Prompt Design for Video Language Models.}\label{prompt}
We use the following prompt for baseline video language models, including EgoVideo~\cite{pei2024egovideo}, InternVideo2~\cite{wang2024internvideo2}, Video-LLaVA~\cite{lin2023video}, VideoLLaMA3~\cite{zhang2025videollama}, GazeGPT~\cite{konrad2024gazegpt}, Voila-A~\cite{yan2024voila}, GazeVQA~\cite{ilaslan2023gazevqa}, and GazeLLM~\cite{rekimoto2025gazellm}:

\textit{``This is an egocentric view video that captures what the driver sees. I am working on a classification task to determine whether the driver is distracted based on the provided video. Distraction is defined as follows: if the driver is looking at an area or object unrelated to the current driving task, they are considered distracted; if the driver is looking at an area or object relevant to the driving task, they are considered attentive. If you determine the driver is inattentive, output 1 directly; if the driver is attentive, output 0 directly.''}

\subsection{Data Annotation Procedure}\label{annotation}

\paragraph{Datasets.}
We use four publicly available driving datasets (DR(eye)VE, BDD-A, DADA-2000, and TrafficGaze) to create our own CogDrive dataset. DR(eye)VE serves as the primary reference, as it is the only existing dataset that provides synchronized egocentric driving videos, eye-gaze signals, and cognitive distraction labels. To enrich the diversity of driving scenarios and data sources, we further incorporate BDD-A, DADA-2000, and TrafficGaze as supplementary datasets. 

Among the four datasets we consider, DR(eye)VE is the only dataset captured from an egocentric perspective with expert-annotated cognitive distraction labels, whereas the other three datasets are collected using forward-facing roof-mounted cameras and lack cognitive annotations. BDD-A focuses on important driving moments sampled from large-scale driving recordings, while DADA-2000 collects segments corresponding to near-crash or crash scenarios; both datasets are captured from roof-mounted vehicle cameras. TrafficGaze is collected under conditions where drivers are explicitly instructed to remain highly attentive, and it also adopts a roof-mounted camera setup. Together, these four datasets cover diverse driving conditions, viewpoints, and attention states, enabling a broader study of cognitive distraction beyond a single dataset.

\begin{figure}[t]
  \centering
    \begin{subfigure}{0.45\linewidth}
      \centering
      \includegraphics[width=\linewidth]{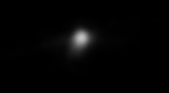}
      \caption{From the whole video}
      \label{density_a}
    \end{subfigure}\hspace{5mm}
    \begin{subfigure}{0.45\linewidth}
      \centering
      \includegraphics[width=\linewidth]{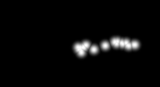}
      \caption{From the video clip}
      \label{density_b}
    \end{subfigure}
  \caption{Examples of fixation density maps used in our annotation protocol.
(a) Fixation density map computed by aggregating eye-gaze over the entire video, resulting in a dominant central area due to long-term averaging.
(b) Fixation density map computed from a short clip of 16 consecutive frames, where multiple gaze peaks may appear due to fine-grained temporal variations.}
  \label{density_map}
\end{figure}

\paragraph{Event Clip Filtering.}
Since DR(eye)VE already provides reliable expert annotations, we directly adopt its labels without re-annotation and use them as the basis for protocol design. Note that the original DR(eye)VE annotations are associated with variable-length segments. For standardization across datasets, we uniformly segment all raw driving videos, including DR(eye)VE, into fixed-length clips of 16 consecutive frames, which serve as the basic units for annotation. For each original video, we compute an average ground-truth fixation density map by aggregating gaze information over the whole video, as illustrated in Fig.~(\ref{density_a}). We also compute a fixation density map for each 16-frame clip, as illustrated in~Fig.~(\ref{density_b}). To identify clips that potentially correspond to distinct cognitive states, we measure the Pearson Correlation Coefficient (CC) between each clip-level density map and the corresponding whole-video average map. Clips with a CC value lower than 0.3 are flagged as candidate event clips, as they indicate gaze patterns that deviate from the overall driving behavior and may correspond to attentive, distracted, or erroneous states.

\paragraph{Annotation Protocol.}
The above candidate clips (with low CC values) are then subjected to manual review and annotation by experts. To ensure consistency and reliability, annotators follow a structured cognitive attention annotation protocol developed with the guidance of domain experts, as summarized in Table~\ref{tab:attention_protocol}. Based on this protocol file, we can get attentive labels and distracted labels. Specifically, we regard some videos as erroneous if there are obvious wrong gaze patterns or wrong video content. We provide this protocol file, named ``cognitive attention protocol.xlsx'', in the supplementary material.

\subsection{Failure Case Analysis}\label{scenario_analysis}

\begin{figure}[t]
  \centering
  \setlength{\tabcolsep}{5pt} %
  \renewcommand{\arraystretch}{1.0} %
  \begin{tabular}{ccc}       
    \begin{subfigure}{0.15\textwidth}
      \centering
      \includegraphics[width=\linewidth]{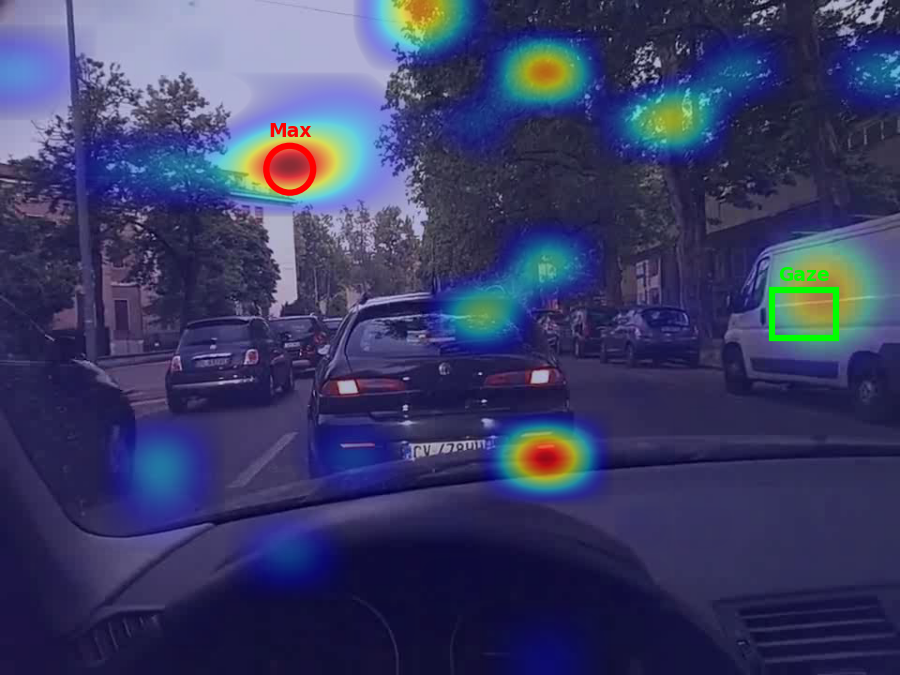}
      \caption{City}
      \label{fig:349}
    \end{subfigure} &
    \begin{subfigure}{0.15\textwidth}
      \centering
      \includegraphics[width=\linewidth]{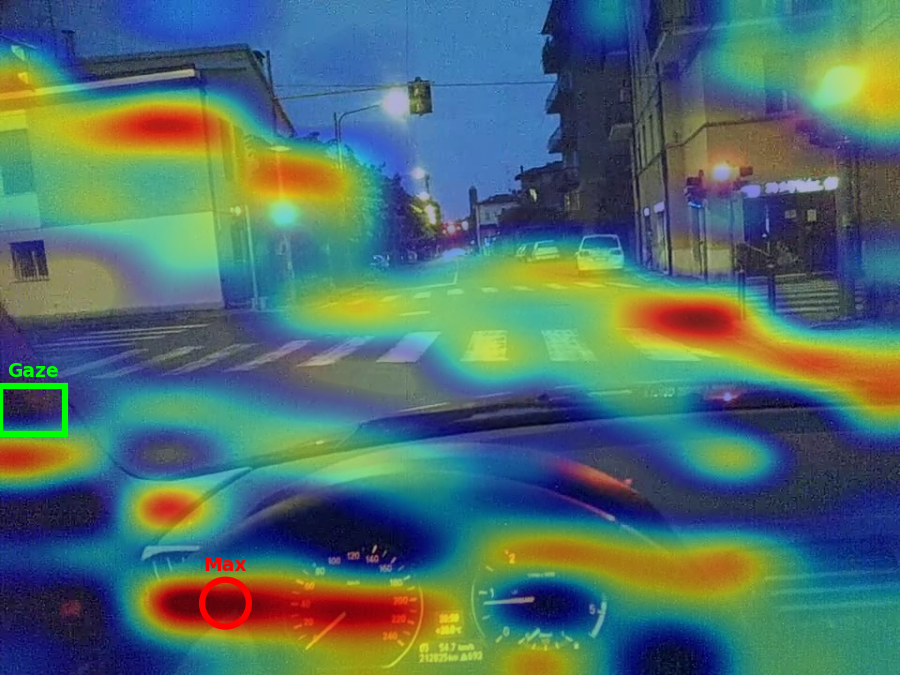}
      \caption{Evening}
      \label{fig:2473}
    \end{subfigure} &
    \begin{subfigure}{0.15\textwidth}
      \centering
      \includegraphics[width=\linewidth]{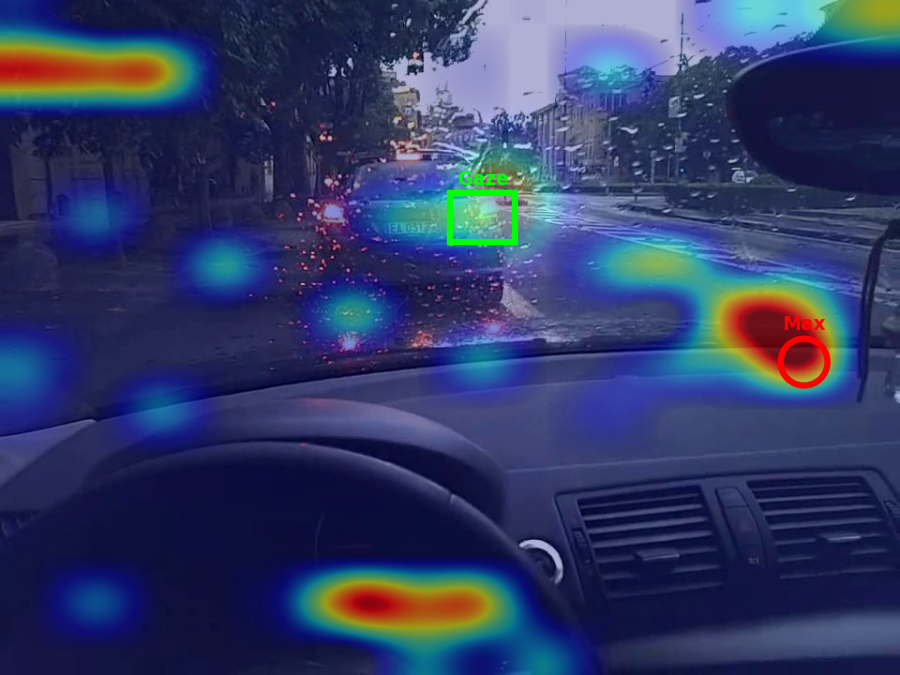}
      \caption{Night}
      \label{fig:306}
    \end{subfigure}
  \end{tabular}
  \caption{Grad-CAM visualizations on representative failure cases. The green box indicates the driver's gaze patch; the red circle marks the peak Grad-CAM activation.}
  \label{fig:scenario_analysis}
\end{figure}

We further analyze representative failure cases using Grad-CAM visualizations, as shown in Fig.~\ref{fig:scenario_analysis}. Ideally, the model should attend to regions that are consistent with the driver's visual attention and relevant to the driving context. However, in challenging scenarios such as city streets, evening illumination, and nighttime driving, we observe clear misalignment between the gaze patch and the most activated visual regions.

This misalignment reveals two main causes of failure. First, the video encoder can be distracted by high-contrast but driving-irrelevant regions, such as reflections, bright lights, or background objects. As a result, the generated video class tokens may encode misleading scene information rather than focusing on safety-critical driving cues. Second, adverse visual conditions, including low illumination, glare, and complex urban backgrounds, can degrade patch-level feature quality. This weakens the gaze-patch alignment in the gaze-guided module and makes it more difficult for the model to associate the driver's fixation with the correct contextual evidence. In addition, \name may fail in complex driving scenes with overlapping objects or visually important but context-irrelevant regions. These observations suggest that improving robustness under challenging visual conditions and enhancing fine-grained gaze-region alignment are important directions for future work.

\subsection{Additional Evaluation Results for \name}\label{more_evaluation}
In Section~\ref{sec:evaluations}, we report the classification accuracy and the F1 score of the distracted class as the primary evaluation metrics. To provide a more comprehensive evaluation of \name, we further include the confusion matrix, Receiver Operating Characteristic (ROC), and Area Under the Curve (AUC) results in this section. These additional evaluation results offer complementary insights into class-wise errors and threshold-independent discrimination performance.

\begin{figure}[t]
    \centering
    \includegraphics[width=0.8\linewidth]{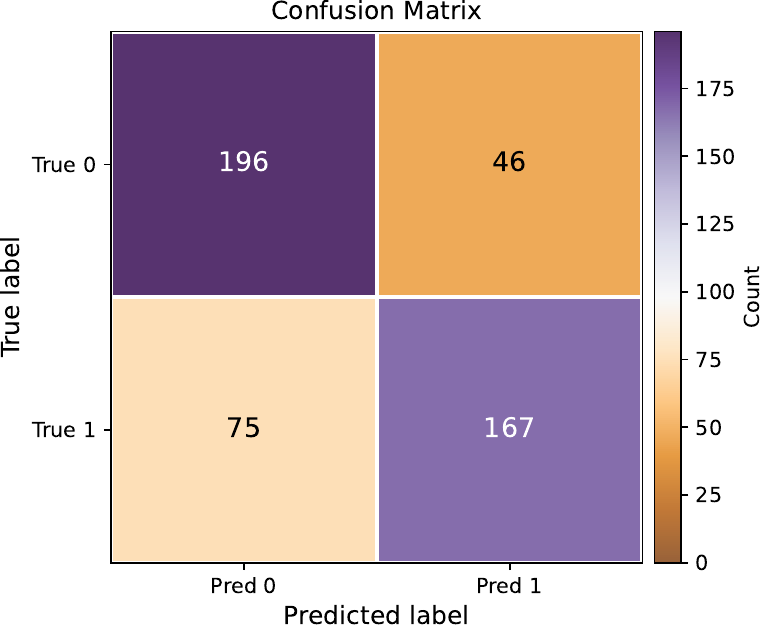}
    \caption{Confusion matrix for \name on the CogDrive dataset.}
    \label{fig:confusion_matrix}
\end{figure}

\paragraph{Confusion Matrix.} 
Fig.~\ref{fig:confusion_matrix} presents the confusion matrix of name on the CogDrive test set, where Class~1 denotes cognitively distracted cases. The matrix is obtained from a single evaluation round, which contains 484 samples, evenly divided into 242 attentive and 242 distracted clips. \name correctly identifies most distracted samples, achieving 167 true positives, while 75 distracted clips are misclassified as attentive. Quantitatively, \name attains a \textit{recall of 0.69} and a \textit{precision of 0.78} for the distracted class, resulting in an \textit{F1 score of 0.74}. These metrics demonstrate a balanced trade-off between sensitivity and reliability in distraction detection.

\begin{figure}[t]
    \centering
    \includegraphics[width=0.65\linewidth]{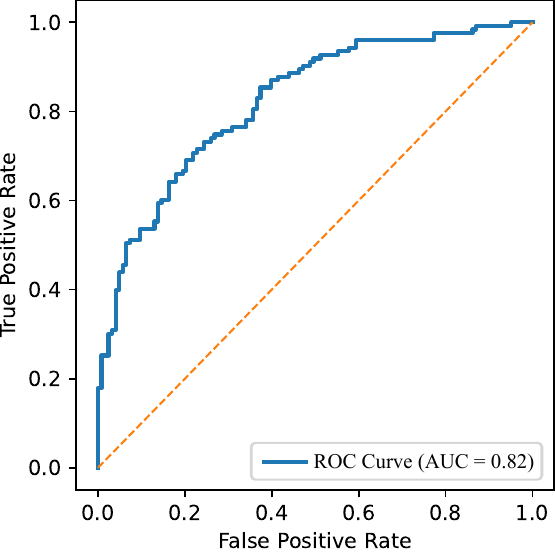}
    \caption{ROC curve and AUC of \name on the CogDrive dataset.}
    \label{fig:roc_auc}
\end{figure}

\paragraph{ROC-AUC Analysis.}
Fig.~\ref{fig:roc_auc} presents the ROC curve, which evaluates the \name's discriminative performance across different decision thresholds by trading off the true positive rate against the false positive rate. The AUC of an ROC curve provides a threshold-independent summary of this capability, where higher values indicate better class separability. \name achieves an AUC of 0.82 on the CogDrive dataset, demonstrating its strong ability to distinguish cognitively distracted cases from attentive ones. The shape of the ROC curve shows that \name maintains a high true positive rate even at low false positive rates, indicating a stable ranking of distracted samples. This suggests that \name effectively captures informative gaze-context interaction patterns rather than relying on a specific classification threshold.

\paragraph{Computational Analysis.}
For computational complexity, we additionally evaluate \name under the TimeSformer backbone setting. In this case, \name contains 138M parameters and requires 760G FLOPs, with an inference speed of 0.4 seconds per sample on an NVIDIA L40S GPU. These results indicate that \name remains computationally practical while achieving efficient inference under a strong video transformer backbone.

\paragraph{Gaze Noise Robustness Analysis.}
To evaluate the robustness of \name to eye-tracking noise and calibration errors, we introduce random perturbations to the gaze coordinates and examine the resulting performance changes. The eye tracker used in our dataset operates at a sampling rate of 30~fps and provides a gaze calibration accuracy of less than $0.5^{\circ}$. When adding 20-pixel random noise to the gaze coordinates, \name achieves an accuracy of 74.17\%, which remains close to the original performance. In contrast, increasing the noise magnitude to 100-pixel reduces the accuracy to 71.07\%. These results suggest that \name is relatively robust to moderate gaze noise, while large deviations from the true fixation region can noticeably degrade performance.

\end{document}